\documentclass[11pt]{article}

\usepackage{pdfcomment}
\usepackage{acl}

\usepackage{times}
\usepackage{latexsym}
\usepackage[T1]{fontenc}
\usepackage[utf8]{inputenc}
\usepackage{microtype}
\usepackage{inconsolata}
\usepackage{graphicx}
\usepackage{booktabs}
\usepackage{amsmath}
\usepackage{amssymb}
\usepackage{enumitem}
\usepackage{xcolor}
\usepackage{tabularx}

\title{A Cost-Aware Agentic Architecture for NL-to-SQL over Nested Enterprise Schemas, with a New Benchmark}

\author{%
Yoga Sri Varshan Varadharajan\thanks{University of Texas at Austin}
\And
Ajay Yadav\footnotemark[1]
\And
Ritesh Goru\thanks{DevRev, USA}
\And
Prateek Chaudhury\thanks{DevRev, Bengaluru, India}
\AND
Constantine Caramanis\footnotemark[1]
\And
Prateek Jain\footnotemark[2]
\And
Divyateja Pasupuleti\footnotemark[3]
\And
Sunil Kumar Pandey\footnotemark[3]
}

\begin{document}
\maketitle

% ============================================================================
\begin{abstract}
% ============================================================================
Natural-language-to-SQL systems have advanced rapidly on academic benchmarks, yet production enterprise schemas exhibit graph-like, semi-structured, deeply nested structure that current benchmarks do not measure. We make two complementary contributions. First, we introduce the DevRev NL2SQL benchmark: 900 execution-verified queries with nested-type and link-graph structure, accompanied by the Semantic Depth Score (SDS), a schema-agnostic rubric for analytical reasoning depth. Second, we present a cost-aware single-generation agentic architecture whose schema-selection, metadata-retrieval, and error-repair components are designed for the requirements this regime imposes. On the DevRev NL2SQL benchmark the system attains 91.7\% answer correctness, a margin of 54.6 percentage points over the next-best baseline; on the Spider~2.0 Snowflake public dataset, it is competitive with leading systems at a single-generation operating point.
\end{abstract}

% ============================================================================
\section{Introduction}
\label{sec:intro}
% ============================================================================

Natural-language interfaces to SQL have advanced rapidly in recent years, with strong results on academic benchmarks including Spider~1.0 \citep{spider1}, BIRD \citep{bird}, and Spider~2.0 \citep{spider2}. 
%Many of the leading systems on the current Spider~2.0 Snowflake leaderboard share a highly effective  design pattern: generate $N$ candidate SQL queries using diverse prompts, decompositions, or multiple models, then select among them via execution-based tournament evaluation or a learned ranker. CHASE-SQL, XiYan-SQL, and Agentar-Scale-SQL are representative of this paradigm. % The approach is highly effective on the current benchmark suite.
% Cut: , and we treat it as an important and active direction for the field.

Deploying NL2SQL on production enterprise data exposes a class of structural challenges that the current benchmark suite does not characterize. Modern enterprise systems of record are often platforms organized around tickets, issues and other workflow objects. These evolve into graph-like, semi-structured data models. Their schemas contain typed \texttt{ARRAY[STRUCT]} columns whose elements have explicit sub-field signatures, polymorphic foreign-key relationships mediated through typed link arrays, and deeply nested objects with multiple levels of field nesting. Analytical questions over such systems require reasoning over array elements, traversing typed link graphs, and producing SQL that combines \texttt{LATERAL~FLATTEN}, conditional aggregation, and enum-constrained filters.

% Earlier version: Existing benchmarks have been invaluable in advancing research on semantic parsing and compositional SQL generation; by design, however, they emphasize flat relational schemas with primitive-typed columns. Spider~2.0-Snow, for example, contains $7{,}860$ tables across $152$ databases, but every table is defined by a flat \texttt{CREATE~TABLE} DDL with primitive-typed columns and the corpus contains essentially no queries that exercise nested-type traversal. The semi-structured, link-graph regime is therefore largely unmeasured by the current benchmark suite.
Existing benchmarks emphasize flat relational schemas with primitive-typed columns. Spider 2.0-Snow, for example, contains 7,860 tables across 152 databases, but every table is defined by a flat CREATE TABLE DDL with primitive-typed columns, only 5\% of tables carry table-level descriptions, and the corpus contains essentially no queries that exercise nested-type traversal. DevRev fields, in contrast, carry English descriptions that document operational intent. The semi-structured, link-graph regime, together with this schema-as-documentation property, is largely unmeasured by the current benchmark suite.

This regime poses important challenges. Graph-like and polymorphic schemas require an iterative, feedback-driven schema-selection mechanism, to better surface the set of tables a query needs. When columns are typed \texttt{ARRAY[STRUCT]}, producing the correct \texttt{LATERAL~FLATTEN} requires the sub-field schema and representative sample values inside the array and not the column type signature alone. Failure modes specific to nested types are diverse and structured (improper \texttt{LATERAL} use, single-element \texttt{UNNEST}, malformed type conversions over nested fields, ambiguous flattening aliases). Generic retry-with-error-message strategies  become suboptimal. These observations motivate the architecture: iterative schema discovery, metadata-rich retrieval, and a structured error model.

Inference cost is a second consideration. Many leading Spider~2.0 systems employ tournament-style candidate selection \citep{chasesql,xiyansql,agentarsql}. On enterprise data with longer schema, longer candidate SQL bodies, slower nested-type executions, and more failure modes that each trigger a retry, the per-query base cost is elevated; this motivates studying single-generation architectures. Our approach could in principle be composed with multi-candidate generation approaches. % that currently dominate the latest benchmarks, a natural direction for future work.

% Cut (pre-announces structure): This paper makes two complementary contributions.

\paragraph{A new benchmark for nested enterprise schemas.}
We introduce and will release the DevRev NL2SQL dataset: 900 execution-verified natural-language queries grounded in a production Snowflake schema with 30 core entity types (190 total schema variants including custom subtypes and snap-in integrations), 13{,}898 fields, 2{,}036 typed \texttt{ARRAY[STRUCT]} columns, 1{,}412 \texttt{STRUCT}-typed or nested fields, and up to four levels of nesting (release details in Appendix~\ref{app:release}). 
%Queries are produced by a four-stage formula-anchored generation pipeline and rewritten across six distinct business-role personas (VP of Sales, Customer Success Manager, Product Manager, Operations Analyst, Finance/RevOps, C-Suite Executive). The corpus exposes three evaluation dimensions that current benchmarks do not measure: nested-field traversal via \texttt{LATERAL~FLATTEN} over \texttt{ARRAY[STRUCT]} columns, polymorphic link-graph reasoning across typed link arrays, and persona-aware lexical disambiguation across business roles. We also introduce the Semantic Depth Score (SDS), a schema-agnostic seven-dimensional rubric for analytical reasoning depth that decouples semantic depth from syntactic clause counts. SDS is used both as a generation control (the pipeline enforces minimum slot scores) and as a diagnostic instrument that places DevRev and Spider~2.0-Snow on a common difficulty axis. Scored on the same 0--5-per-dimension rubric (maximum 35), the DevRev corpus achieves a mean SDS of 19.15, comparable to Spider~2.0-Snow's 20.20, with 90.4\% of DevRev queries at SDS~$\geq 11$.
% Cut trailing: , confirming the dataset occupies the high end of the analytical difficulty spectrum.

\paragraph{A cost-aware single-generation agentic architecture.}
The system has five components. An LLM-driven schema-discovery loop browses a precomputed knowledge graph through tool calls and re-invokes those tools on schema-coverage errors, broadening context in response to execution feedback. 
%The per-table schemas it returns use a compressed one-line-per-field format that exposes field types, enum vocabularies, English descriptions, and sub-field signatures for \texttt{ARRAY[STRUCT]} and \texttt{STRUCT} columns. 
A structured error taxonomy of 15+ classes maps each execution failure to a targeted repair directive.% and injects the full history of prior failed attempts into the next generation call. 
A pre-execution query plan verifier audits the SQL for structural flaws. % (e.g., Cartesian products, incorrect grain) before execution. 
A deterministic checker chain combines static checks with rewriters that auto-fix cross-dialect functions and identifier casing. A dynamic cheatsheet \citep{dynamiccheatsheet} accumulates transferable SQL rules across queries. % and injects them into the generator (test-time learning). 
A semantic validator inspects executed results against the original question and rejects clear contradictions. The system uses a single generation trajectory per query and does not require fine-tuning. On the DevRev benchmark our system attains 91.7\%  correctness on the full 900-query corpus and 91.5\% on the high-SDS subset (SDS~$\geq 11$); on the Spider~2.0 Snowflake public leaderboard it is competitive with leading systems at a single-generation operating point.
% Earlier version: The system has four components. A four-stage schema-selection pipeline progressively refines context, with a final error-guided expansion stage that fetches additional tables when execution feedback reveals link-graph under-selection. A metadata-enriched retrieval stage surfaces column profiles, sample values, sub-field schemas, and join hints. A structured error taxonomy of 15+ classes maps each execution failure to a targeted repair directive and injects the full history of prior failed attempts into the next generation call. A deterministic checker tool-chain catches common SQL anti-patterns before any database execution. The system uses a single generation trajectory per query and does not require fine-tuning.

% Cut (pure recap): We view these contributions as complementary to existing work: the benchmark extends evaluation into a regime that prior datasets did not target, and the architecture provides a single-generation operating point on the cost--accuracy frontier that does not preclude composition with multi-candidate selection.

% ============================================================================
\section{Related Work}
\label{sec:related}
% ============================================================================

%Our work intersects several active threads in NL2SQL research: multi-candidate test-time scaling, agentic and tool-augmented generation, schema linking, error-guided self-repair, and benchmark construction.

%\subsection{Multi-Candidate and Test-Time Scaling}

{\bf Multi-Candidate and Test-Time Scaling.} Tournament-style selection has become a dominant NL2SQL paradigm. CHASE-SQL \citep{chasesql} combines multi-path reasoning with preference-optimized candidate selection. XiYan-SQL \citep{xiyansql} runs multiple generators with diverse prompting strategies and ensembles their outputs. Agentar-Scale-SQL \citep{agentarsql} scales test-time compute through parallel synthesis followed by sequential refinement. DeepEye-SQL \citep{deepeyesql} pairs $N$-version SQL generation with execution-guided confidence-aware selection. % inside a software-development-inspired pipeline. These approaches trade inference cost for accuracy at test time, typically at 5--20$\times$ the compute of a single-generation system. Our architecture occupies a different operating point on the cost--accuracy frontier; the design principles in Section~\ref{sec:principles} are largely orthogonal to multi-candidate selection and could in principle be composed with it.

%\subsection{Agentic and Tool-Augmented NL2SQL}
{\bf Agentic and Tool-Augmented NL2SQL.}
A complementary line of work frames NL2SQL as a multi-step tool-calling process. % in which the model interacts with tools such as schema lookups, executions, error inspections, rather than producing SQL in one shot. 
MAC-SQL \citep{macsql} introduces multi-agent collaboration  with a decomposer, refiner, and selector. DIN-SQL \citep{dinsql} decomposes the task into sub-problems and applies in-context learning with self-correction. DAIL-SQL \citep{dailsql} systematizes the design space of LLM-based generation strategies. AskData \citep{askdata} targets enterprise NL2SQL through offline metadata extraction that enriches schema context. Our two-tier orchestrator design (Section~\ref{sec:two-tier}) follows this agentic flow, separating workflow control from SQL synthesis; the contribution is in how the workflow is structured around the regime properties characterized in Section~\ref{sec:benchmark}: in particular, the iterative schema-expansion loop and the structured error taxonomy that drives it.

{\bf NL2SQL Benchmarks.}
Spider~1.0 \citep{spider1} established the cross-domain NL2SQL evaluation paradigm with 200 databases of moderate complexity and primitive-typed columns. BIRD introduced messier schemas and external-knowledge questions but retained the flat-table assumption. Spider~2.0 \citep{spider2} expanded to production-scale schemas; its Snowflake subset (Spider~2.0-Snow) contains 7{,}860 tables across 152 databases and shifts the schema-linking challenge to large scale. Its tables remain flat in the DDL sense: no declared sub-field schemas, no typed link arrays. The DevRev benchmark we introduce in Section~\ref{sec:benchmark} complements Spider~2.0-Snow by stressing nested-type traversal and link-graph reasoning on one production schema; the two settings exercise different aspects of enterprise NL2SQL, and together provide better coverage of the design space than either alone.

% \vspace{-5pt}
{\bf Synthetic NL query generation.}
SING-SQL \citep{singsql} is a concurrent automated framework for generating in-domain NL2SQL training data. It hierarchically partitions a database into sub-schemas.
%, first selecting joinable table subsets, then applying a sliding-window strategy over non-key columns, 
and synthesizes SQL across multiple complexity tiers.
%, with LLM-as-judge validation and automatic repair. 
Our DevRev generation pipeline % shares the motivation of controlled sub-schema construction to ensure coverage and analytical depth, but 
differs in: (1) we target evaluation queries rather than fine-tuning data; (2) we use our SDS dimension targets, prioritizing analytical reasoning depth; and (3) we apply six-persona rewriting to introduce lexical variation across business roles. Finally, SING-SQL  operates on flat schemas, while DevRev is designed specifically for the nested-type, link-graph regime.

% ============================================================================
\section{The DevRev Benchmark}
\label{sec:benchmark}
% ============================================================================

\subsection{The Enterprise Schema Regime}
\label{sec:regime}

Existing NL2SQL benchmarks evaluate systems on flat, tabular schemas with primitive-typed columns. Spider~2.0-Snow spans 547 queries over 152 databases and 7{,}860 tables. Every table is defined by a flat \texttt{CREATE TABLE} DDL; the corpus contains no \texttt{STRUCT} columns, and where semi-structured data appears the columns are typed as opaque \texttt{VARIANT} with no sub-field schema in the DDL.

Production enterprise schemas are structurally different. Work-item platforms (organized around tickets, issues, and other workflow objects) are graph-like: core entities are nodes and typed link arrays, tags, and group memberships are edges. Their schemas contain \texttt{ARRAY[STRUCT]} columns with explicit sub-field signatures (not opaque \texttt{VARIANT}), polymorphic relationships mediated through typed link arrays whose \texttt{type} sub-field discriminates the linked entity, objects nested up to four levels deep, and per-field English descriptions that encode operational intent (e.g., whether an \texttt{INTEGER} column is a count or a boolean flag, or what the labels in a \texttt{stage} enum denote). Analytical questions over such schemas require \texttt{LATERAL~FLATTEN} over array columns, enum-constrained filtering over embedded vocab, and multi-hop typed-link traversal patterns absent from the existing benchmark suite.

Table~\ref{tab:schema_compare} summarizes the contrast: Spider~2.0-Snow exercises large-scale schema linking over many flat tables; DevRev exercises nested-type traversal and link-graph reasoning over one production schema.

\begin{table}[t]
\centering
\small
\begin{tabular}{@{}lrr@{}}
\toprule
\textbf{Property} & \textbf{DevRev} & \textbf{Spider~2.0-Snow} \\
\midrule
Databases / domains       & 1            & 152 \\
Tables                    & 15$^\dagger$ & 7{,}860 \\
\midrule
Typed \texttt{ARRAY[STRUCT]} fields & 2{,}036 & 0 \\
\texttt{STRUCT} / nested fields     & 1{,}412 & 0 \\
Opaque \texttt{VARIANT} columns     & 0       & 6{,}180 \\
Max nesting depth                   & 4       & 1 \\
\bottomrule
\end{tabular}
\caption{Schema comparison. DevRev's \texttt{ARRAY[STRUCT]} fields have explicit sub-field schemas; Spider~2.0-Snow's \texttt{VARIANT} columns are
opaque semi-structured types with no DDL sub-field definition.
$^\dagger$15 of 30 core entity types (190 total schema variants); 15 used for benchmark generation.}
\label{tab:schema_compare}
\end{table}

\subsection{Schema Overview}
\label{sec:schema}

The DevRev production database contains 30 core entity types and 190 schema variants; our benchmark uses the 15 for which detailed m-schema documentation was available. The schema counts appear in Table~\ref{tab:schema_compare}.

Two structural patterns dominate. First, the \emph{polymorphic link graph}: a ticket's \texttt{links} field is an \texttt{ARRAY[STRUCT]} whose elements carry a \texttt{type} sub-field (\texttt{blocks}, \texttt{relates\_to}, \texttt{resolves}, \ldots) and a \texttt{display\_id} pointing at the linked entity. Because the same array can link to issues, enhancements, or other tickets, resolving the join target requires filtering on \texttt{type} and consulting an external entity mapping a pattern with no analogue in flat-table benchmarks. Second, \emph{non-link \texttt{ARRAY[STRUCT]} fields}: \texttt{surveys\_aggregation} holds survey-response records, \texttt{tags} holds label structs with type and source attribution, and \texttt{owned\_by} holds typed user references. Each requires \texttt{LATERAL~FLATTEN} to access, with a distinct sub-field signature per entity.

\subsection{Four-Stage Generation Pipeline}
\label{sec:pipeline}

Queries are generated by a four-stage pipeline enforcing analytical depth: (1) Column Planning selects column subsets; (2) Formula Retrieval derives structural patterns; (3) Query Generation generates natural language queries using DevRev-specific column names and business concepts; and (4) Persona Rewriting rewrites each slot query into a distinct business role voice. We provide the complete details in the Appendix.

\subsection{The Semantic Depth Score (SDS)}
\label{sec:sds}

SDS is a seven-dimensional rubric that measures the \emph{analytical reasoning} required to formulate a correct SQL query, independent of the resulting query's syntactic complexity. Each dimension is scored 0--5 (max of 35). The dimensions are:

\begin{itemize}[noitemsep,leftmargin=*]
  \item \textbf{D1 Eligibility:} how many intersecting conditions define the subject of analysis.
  \item \textbf{D2 Derived Metric:} whether a non-trivial value (ratio, delta, unit conversion) must be computed rather than read directly.
  \item \textbf{D3 Population Scoping:} how tightly the analysis population is restricted by independent filters.
  \item \textbf{D4 Multi-Stage Logic:} whether multiple analytical steps are chained (find winner, then compute for winner only).
  \item \textbf{D5 Output Precision:} how many distinct output fields are explicitly named or implied.
  \item \textbf{D6 Temporal Precision:} whether a precise time window, date, or rolling period is required.
  \item \textbf{D7 SQL Construct Diversity:} how many SQL constructs (window functions, CTEs, \texttt{LATERAL~FLATTEN}, set operations) are needed.
\end{itemize}

SDS decouples semantic depth from syntactic clause counts, which is a limitation of the easy/medium/hard/extra-hard labels used by earlier benchmarks: a query with a single deeply chained derivation can score high on D1, D2, and D4 while using few clauses, and a query with many simple aggregations can score low on D2 and D4 despite producing long SQL. SDS supports diagnostic analysis: a system can be evaluated separately on high-D6 queries (temporal-reasoning failures) versus high-D7 queries (advanced-construct failures) to identify targeted improvement opportunities.

\section{System Design for Nested Enterprise Schemas}
\label{sec:method}
% ============================================================================

\begin{figure*}[t]
\centering
\includegraphics[width=\textwidth]{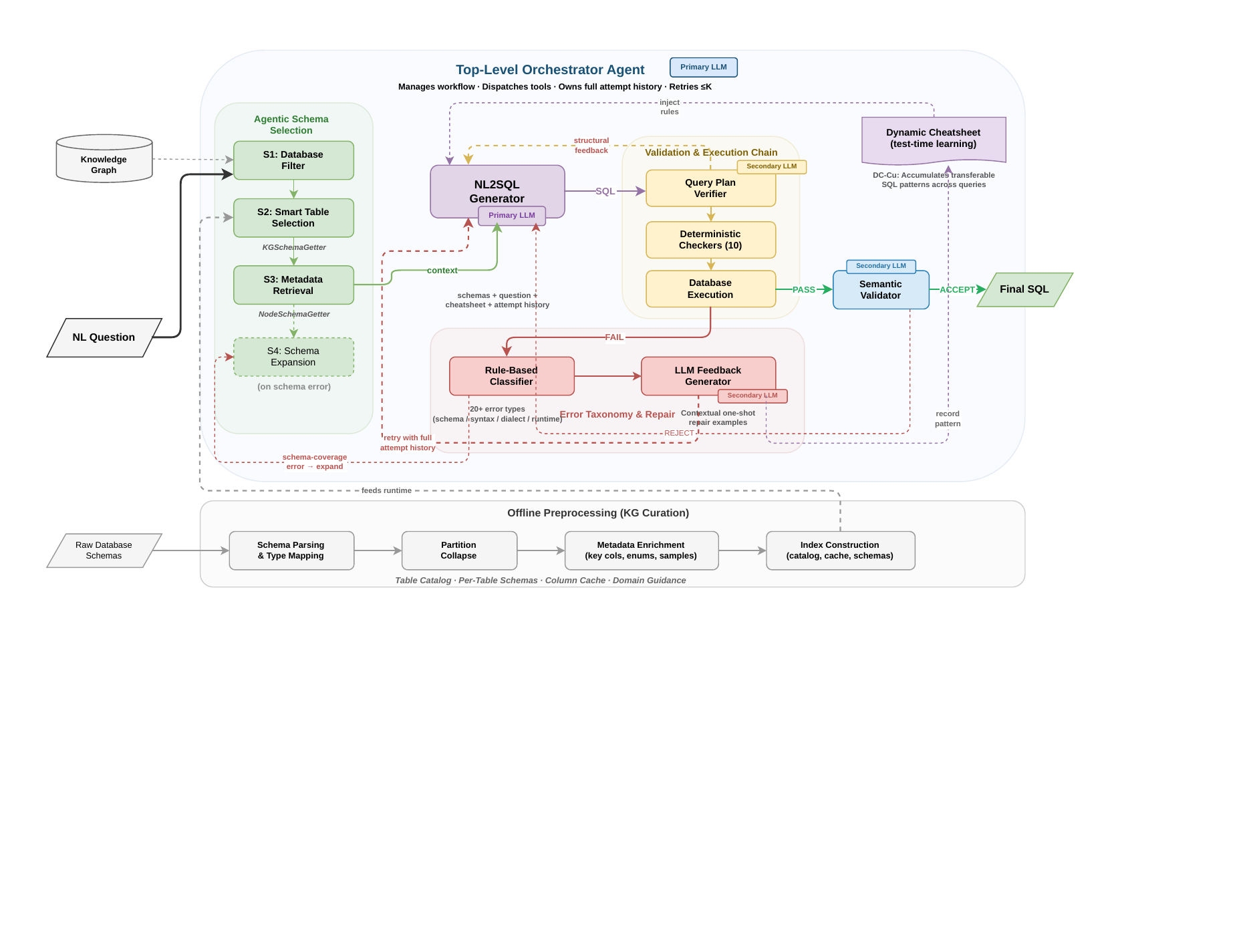}
\vspace{-150pt}
\caption{System architecture. The top-level agent manages the workflow by invoking three tools: \texttt{KGSchemaGetter} (catalog browsing), \texttt{NodeSchemaGetter} (compressed per-table schema retrieval), and \texttt{NLToSQL} (SQL generation).}
\label{fig:architecture}
\end{figure*}

The design principles derive from the schema regime characterized in Section~\ref{sec:benchmark}. Figure~\ref{fig:architecture} shows the architecture.

\subsection{Design Principles from the Regime}
\label{sec:principles}

The architecture is organized around four regime-inspired principles:
\textbf{(1) Iterative schema selection.} Polymorphic typed-link arrays prevent the required table set from being predictable before generation, so the system must expand schema context in response to execution feedback rather than committing to a fixed set up front.
\textbf{(2) Sub-field-aware retrieval.} For \texttt{ARRAY[STRUCT]} columns, DDL type signatures are insufficient; the generator needs nested sub-field schemas, enum vocabularies, and sample values.
\textbf{(3) Structured, history-aware error handling.} Nested-type failure modes are diverse yet classifiable; a taxonomy with targeted repair directives and full attempt history outperforms generic retry-with-raw-error.
\textbf{(4) Single-generation operating point.} Elevated per-query cost on enterprise schemas shifts the tradeoff away from multi-candidate tournaments; principles 1--3 reduce the variance that $N$-candidate generation would otherwise absorb.

\subsection{Two-Tier Agentic Architecture}
\label{sec:two-tier}

The system separates workflow orchestration from SQL synthesis. A \textbf{top-level orchestrator agent} receives the natural-language question and manages the full query lifecycle by invoking three tools: \texttt{KGSchemaGetter} (catalog browsing), \texttt{NodeSchemaGetter} (per-table schema retrieval), and \texttt{NLToSQL} (SQL generation). It inspects execution results, classifies errors, and decides whether to expand schema context, repair the query, or accept the result. The orchestrator is the only component that holds the full attempt history and interacts with the database.

The \textbf{NL2SQL generator} is a tool call, not a separate agent: it takes assembled context (schemas, question, cheatsheet rules, structured feedback from prior failures) and returns a single SQL query. It does not call tools or see execution results.% directly.

This separation keeps each component's prompt focused: meta-level decisions (e.g., ``fetch additional tables after a \texttt{table\_not\_found} error'') belong to the orchestrator and do not contaminate the generator's SQL synthesis task. The orchestrator and generator use a frontier model for deep reasoning; constrained verification tasks (Query Plan Verifier, Semantic Validator, error feedback generation) are routed to a faster, cheaper model, keeping the iterative repair loop cost-efficient.

\subsection{LLM-Driven Iterative Schema Discovery}
\label{sec:schema-pipeline}

Schema selection is driven by the orchestrator LLM through tool calls against a precomputed knowledge graph. There is no embedding-based ranker in the runtime hot path. The LLM itself selects which tables to expose to the generator and revises that selection in response to execution feedback. The knowledge graph is built offline from source schema documentation via three transformations: partition collapsing (reducing millions of date-partitioned physical tables to thousands of logical entries), key-column scoring (surfacing the most informative columns in the catalog), and type normalization (mapping source-specific types to a canonical vocabulary). Details are in Appendix~\ref{app:offline-kg}.

\noindent
{\bf Stage 1: Catalog browsing.}
The orchestrator calls \texttt{KGSchemaGetter(database\_name)} to retrieve the curated table catalog. Each entry contains the table name, a one-line description, and the table's top key columns from offline scoring.

\noindent
{\bf Stage 2: Smart table selection in context.}
The orchestrator reads the catalog inline and decides which tables are relevant to the question. Selection is LLM-driven and conditioned on the question, the catalog (which includes top-K key columns to aid selection), and the cheatsheet (Section~\ref{sec:cheatsheet}).

\noindent
{\bf Stage 3: Per-table schema retrieval.}
The orchestrator calls \texttt{NodeSchemaGetter(table\_id)} in parallel for each chosen table. Each call returns a compressed schema in a one-line-per-field \emph{Schemonic} format that captures the field name, type, enum vocabulary, English description, and sub-field schema for nested columns. The compression is essential because a single DevRev entity (e.g., \texttt{ticket}) has hundreds of fields and would otherwise dominate the context budget.

\noindent
{\bf Stage 4: Error-guided schema expansion.}
When generated SQL fails with a schema-coverage error (\texttt{table\_not\_found}, \texttt{column\_not\_found}, \texttt{unresolved\_alias}), the orchestrator re-invokes \texttt{KGSchemaGetter} or \texttt{NodeSchemaGetter} for additional tables, guided by the identifiers the failed SQL referenced, and reassembles the context. This makes schema discovery iterative, handling the polymorphic-link case (Section~\ref{sec:schema}) where the required table set cannot be predicted before generation.

\subsection{Pre-Execution Query Plan Verifier}
\label{sec:plan-verifier}

Before a generated query is passed to the deterministic checkers or executed against the database, it is audited by an LLM-based Query Plan Verifier. This component acts as a structural linter, analyzing the SQL for logical flaws that would otherwise result in silent semantic failures or expensive execution errors. The verifier checks for three specific conditions:

\begin{itemize}[noitemsep,leftmargin=*]
  \item \textbf{Join Safety:} Ensures that joins are performed on valid key columns rather than arbitrary fields, preventing accidental Cartesian products.
  \item \textbf{Grain Correctness:} Verifies that the level of detail (grain) of the query matches the user's question, particularly when aggregating over flattened arrays or joined tables.
  \item \textbf{Deduplication Placement:} Checks that \texttt{DISTINCT} or \texttt{GROUP BY} clauses are correctly positioned to handle fan-out from one-to-many joins or array unnesting.
\end{itemize}

If the verifier detects a structural flaw, it rejects the query and provides targeted feedback (e.g., ``The join on \texttt{status} will cause a Cartesian product; join on \texttt{id} instead''). This feedback is immediately routed back to the generator, bypassing the database entirely. By catching logical errors pre-execution, the verifier reduces both database compute costs and the incidence of plausible-looking but incorrect result sets.

\subsection{Dynamic Error Taxonomy and History-Aware Feedback}
\label{sec:error-taxonomy}

When generation produces SQL that fails on execution, the orchestrator classifies the failure into one of 15+ error types organized along two axes: failure phase (schema coverage, nested-type, syntax/dialect, runtime) and corrective action. The full taxonomy is in Appendix~\ref{app:error-taxonomy}. The key routing decision is: schema-coverage errors (\texttt{table\_not\_found}, \texttt{column\_not\_found}, \texttt{unresolved\_alias}) trigger Stage~4 schema expansion, addressing the root cause (missing context) rather than the symptom; all other error types trigger in-place repair with a targeted directive.

% \vspace{-5pt}
Each error class maps to a concise repair directive.
%describing what went wrong and how to fix it. 
Together with the full history of prior failed attempts, this is injected as structured context into the next generator call. This prevents the generator from repeating classified mistakes and enables progressive refinement across retries.

The taxonomy is dynamic: error definitions and repair directives live in a configuration file that can be extended without code changes. Section~\ref{sec:ablation} evaluates the contribution of structured, history-aware feedback against a generic retry-with-error-message baseline.

\subsection{Deterministic Checker Chain}
\label{sec:checkers}

Generated SQL passes through a chain of deterministic stages (summarized in Table~\ref{tab:checker-chain} in the Appendix) before the orchestrator accepts a result. The chain mixes two kinds of stages. \emph{Blockers} are pure functions that either pass the SQL through or return a classified error that re-enters the repair loop. \emph{Rewriters} silently transform the SQL and hand the rewritten form to the next stage; they never fail.

%\begin{table}[t]
%\centering
%\scriptsize
%\begin{tabularx}{\columnwidth}{@{}llX@{}}
%\toprule
%\textbf{Checker} & \textbf{Type} & \textbf{Purpose} \\
%\midrule
%Multi-Statement & Blocker & Rejects multiple statements \\
%\texttt{SELECT *} Guard & Blocker & Rejects \texttt{SELECT *} patterns \\
%\texttt{VALUES} Guard & Blocker & Rejects \texttt{VALUES(...)} in CTEs \\
%Placeholder Guard & Blocker & Rejects bind parameters/stubs \\
%Dialect Rewriter & Rewriter & Silently fixes dialect mistakes %\\
%Reserved Word Guard & Blocker & Rejects reserved word aliases %\\
%Syntax Validator & Blocker & Local parse check %(\texttt{sqlglot}) \\
%Fuzzy Search Guard & Blocker & Rejects %\texttt{LIKE}/\texttt{ILIKE} patterns \\
%Heuristic Detector & Blocker & Regex checks for anti-patterns \\
%Execution Validator & Blocker & Executes against the database \\
%\bottomrule
%\end{tabularx}
%\caption{Deterministic Checker Chain. Blockers reject queries %and return errors to the orchestrator; Rewriters silently %mutate the SQL.}
%\label{tab:checker-chain}
%\end{table}
% \vspace{-5pt}

Blockers prevent \emph{silent semantic failures} (queries that execute successfully but return incorrect results, e.g., due to overly permissive fuzzy matching on enum columns). Rewriters silently fix cross-dialect slippage: a Column Identifier Fixer normalizes casing, and a Dialect Auto-Rewriter maps BigQuery/DuckDB functions to their Snowflake equivalents (e.g., \texttt{DATE\_DIFF}~$\to$~\texttt{DATEDIFF}). Execution is the chain's final stage; cheap static checks run first so that detectable failures are absorbed before incurring database cost.

\subsection{Test-Time Learning}
\label{sec:cheatsheet}

We use the \emph{Dynamic Cheatsheet} framework of \citet{dynamiccheatsheet}: a persistent, evolving memory of patterns curated at test time. In our setting the entries are SQL rules. When the orchestrator classifies an execution error, it extracts a concise, transferable rule (e.g., ``divide THELOOK timestamps by 1000000 before casting'' or ``use \texttt{LATERAL FLATTEN} for the \texttt{cpc} array''). These rules are persisted across the evaluation session. Before every generation call, the top rules from the cheatsheet are injected into the generator's prompt.
% Cut trailing: This test-time learning mechanism ensures that once the system discovers a domain-specific convention or a strict dialect requirement through trial and error, it does not repeat the mistake on subsequent queries.

\subsection{Semantic Validation}
\label{sec:semantic-validation}

After a generated query successfully executes against the database, the system performs a final semantic plausibility check. An LLM validator is presented with the user's original question, the generated SQL, and a preview of the execution results (column names and the first few rows). The validator is instructed to be lenient (accepting empty result sets if the filters are restrictive, and accepting any result that is directionally plausible) but to reject clear contradictions. For example, if the user asks for ``how many tickets'' and the result set contains hundreds of raw ticket rows instead of a single aggregate count, the validator rejects the result. Rejections are treated as semantic errors: the validator's explanation is appended to the attempt history, and the generator is re-invoked.

% ============================================================================
\section{Experiments}
\label{sec:results}
% ============================================================================
The complete description is in the Appendix. 

%We evaluate the system on two tracks. The first is the DevRev benchmark from Section~\ref{sec:benchmark}, the regime the architecture is designed for. The second is the Spider~2.0 Snowflake leaderboard, a flat-table, large-scale schema-linking regime that the leading multi-candidate systems were optimized on. The first track tests whether the design principles in Section~\ref{sec:principles} translate into accuracy on the regime they target. The second tests whether the architecture is competitive at a single-generation operating point on the regime where multi-candidate generation currently dominates.

\subsection{Experimental Setup}
\label{sec:setup}
{\bf Baselines.} We compare against \textbf{APEX-SQL}~\cite{cao2026apex} (Hypothesis-Verification loop, run with \texttt{num\_votes=1} to match our single-path compute budget), \textbf{FlexSQL}~\cite{pham2026flexsql} (plan-and-stitch with four candidate plans, default configuration), and \textbf{ReFoRCE}~\cite{deng2025reforce} (self-refinement with majority-vote consensus). %All baselines are run on the DevRev schema without modification. %APEX-SQL's default \texttt{num\_votes=5} would consume $5\times$ more inference than any other system; we use \texttt{num\_votes=1} for a cost-normalized comparison.

{\bf Models.} All four systems use \texttt{gpt-5.2-2025-12-11} as the primary model. Our system additionally routes verification and feedback tasks (query plan verifier, semantic validator) to the secondary model -- \texttt{gpt-4.1-mini} to reduce cost. No fine-tuning is performed.

{\bf Metrics.} Our primary metric is \textbf{Answer Correctness (AC)}: the final verdict of a multi-stage evaluation pipeline. Generated SQL must first pass the deterministic checker chain (10 static/rewrite stages), then execute successfully against the live database, and finally pass a semantic LLM validation (GPT-5.2 judge) that receives the NL query, generated SQL, and a preview of result rows and issues a binary verdict on whether the result answers the question. AC is therefore strictly stronger than Execution Accuracy (EX): it requires syntactic validity, successful execution, \emph{and} semantic correctness. We additionally report \textbf{Mean Score (\%)}, the continuous judge score normalised to 0--100\% of a 50-point rubric, as a secondary quality indicator. Full judge criteria are in Appendix~\ref{app:judge}. For Spider~2.0-Snow we report EX using the official evaluation script provided by \citet{spider2}.

%{\bf Cost accounting.} Wall-clock time and token consumption are measured on a stratified 50-query subsample using the same model for all systems.

\subsection{Main Results}
\label{sec:main_results}

\begin{table*}[t]
\centering\small
\setlength{\tabcolsep}{5pt}
\begin{tabular*}{\textwidth}{@{\extracolsep{\fill}}lcccccc@{}}
\toprule
 & \multicolumn{3}{c}{\textbf{Accuracy}}
 & \multicolumn{3}{c}{\textbf{Cost \& Latency}} \\
\cmidrule(lr){2-4}\cmidrule(l){5-7}
\textbf{System}
  & \textbf{DevRev AC (\%)} & \textbf{DevRev Mean (\%)} & \textbf{Spider 2.0-Snow EX (\%)}
  & \textbf{Wall (s/q)} & \textbf{Tokens (K/q)} & \textbf{\$/correct q} \\
\midrule
Ours     & \textbf{91.7} & 79.2 & 85.8 (103/120)$^\ddagger$  & 161     & 108     & \textbf{\$0.57} \\
APEX-SQL & 27.2          & 47.2 & 73.13$^\S$                 & 200     & 260     & \$4.60          \\
FlexSQL  & 37.1          & 46.4 & 65.45$^\S$                 & 1{,}653 & 1{,}183 & \$15.35         \\
ReFoRCE  & 29.1          & 33.0 & 62.89$^\S$                 & 74      & 56      & \$0.93          \\
\bottomrule
\end{tabular*}
\caption{Results on DevRev-900 and Spider~2.0-Snow. \textbf{\$/correct~q} = API cost per correctly answered query (gpt-5.2 pricing: \$1.75/M input, \$14.00/M output).
$^\ddagger$120-query gold-answer subset; not comparable to full leaderboard.
$^\S$Best published leaderboard submission per system.}
\label{tab:main_results}
\end{table*}

Table~\ref{tab:main_results} reports results on the full 900-query DevRev benchmark. Our system achieves Answer Correctness of 91.7\% (825/900 queries), a margin of \textbf{54.6~pp} over the next-best system (FlexSQL, 37.1\%). The baselines cluster between 27--37\% despite very different architectures, hypothesis-verification loop (APEX-SQL), plan-and-stitch (FlexSQL), and majority-vote consensus (ReFoRCE), suggesting all three hit the same structural ceiling: the inability to reason over nested \texttt{ARRAY[STRUCT]} fields and typed link traversal, rather than any flaw intrinsic to their respective generation strategies.

% ReFoRCE is instructive: its majority-vote mechanism scores lower than FlexSQL (29.1\% vs.\ 37.1\%) and 507 of 900 queries produced no SQL at all, revealing that on enterprise schemas failure is systematic: voting amplifies errors rather than filtering them. The cost columns reinforce this: FlexSQL consumes 21$\times$ more tokens than ReFoRCE and takes over 27~minutes per query, yet answers fewer than 2-in-5 queries correctly.

The \$/correct~q column makes the production deployment case directly. Using verified \texttt{gpt-5.2-2025-12-11} pricing (\$1.75/M input, \$14.00/M output), our system costs \textbf{\$0.57 per correctly answered query}: 8$\times$ cheaper than APEX-SQL (\$4.60), 27$\times$ cheaper than FlexSQL (\$15.35), and 1.6$\times$ cheaper than ReFoRCE (\$0.93). Notably, ReFoRCE is the cheapest system per query (\$0.27) but only 29.1\% of its queries are correct, so the effective cost per useful answer is 1.6$\times$ ours.
\subsection{Difficulty-Stratified Analysis}
\label{sec:sds_analysis}
% ============================================================================

\begin{figure}[t]
  \centering
  \includegraphics[width=\columnwidth]{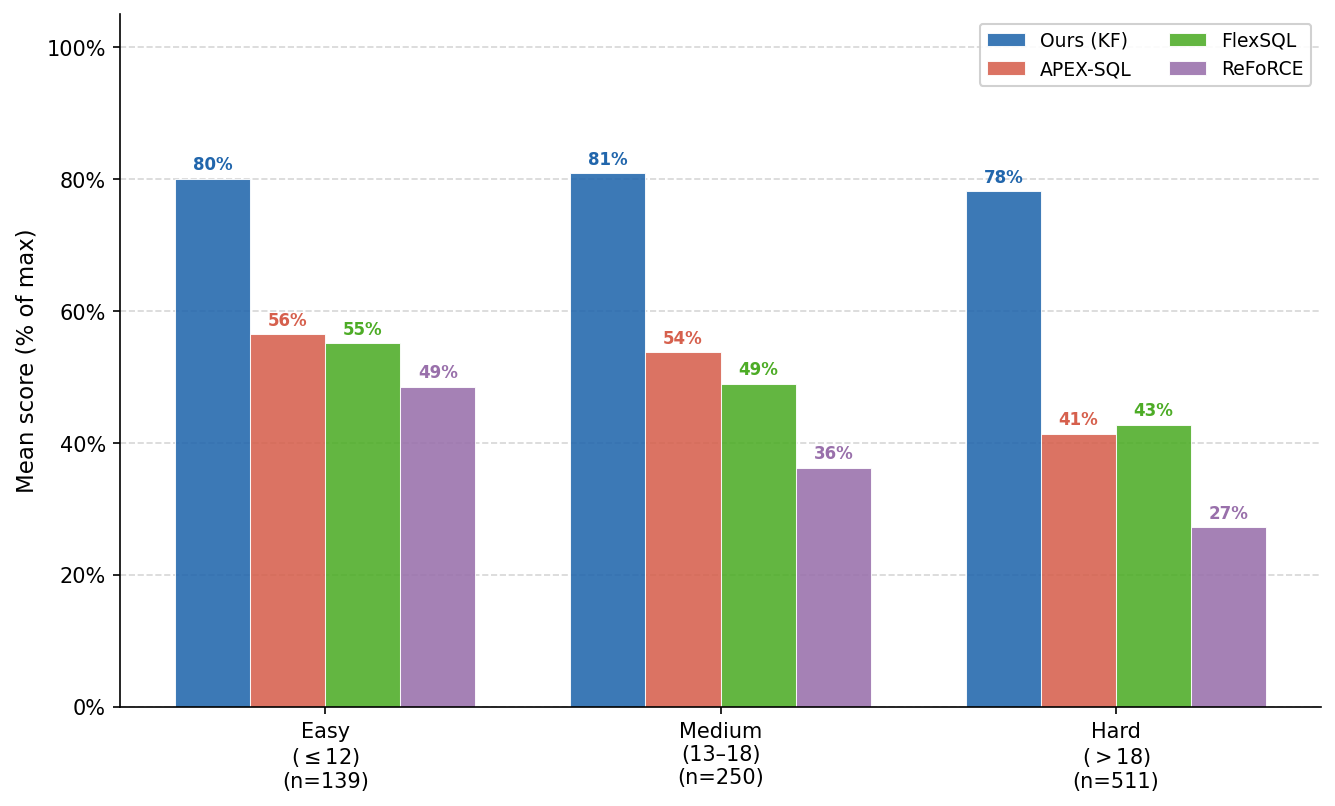}
  \caption{Mean judge score (\% of 50-point maximum) by SDS bucket for all four
  systems.}
  \label{fig:sds_results}
\end{figure}

Figure~\ref{fig:sds_results} stratifies results by SDS total into three buckets: Easy ($\leq$12, $n$=139), Medium (13--18, $n$=250), and Hard ($>$18, $n$=511), and reveals two different responses to increasing complexity.

All three baselines degrade monotonically: ReFoRCE most steeply ($-$21.3~pp, from 48.5\% to 27.2\%), followed by APEX-SQL ($-$15.1~pp) and FlexSQL ($-$12.3~pp). The full per-bucket breakdown is in Appendix~\ref{app:sds_breakdown}. Our system's mean score remains essentially flat across buckets (80.1\% $\to$ 80.9\% $\to$ 78.1\%, a 2.8~pp range), and Answer Correctness similarly spans less than 6~pp (89.2\% $\to$ 94.4\% $\to$ 91.0\%). As a result, the gap over FlexSQL \emph{grows} with complexity: 44.6~pp on Easy, 54.8~pp on Medium, and 57.1~pp on Hard.

% ============================================================================
\subsection{SDS as an Actionable Improvement Signal}
\label{sec:sds_actionable}
% ============================================================================

SDS as a diagnostic can  improve agent performance without modifying core architecture. %We validate this on two baselines by providing SDS-aware hints to their prompts on the queries where they originally failed.

\paragraph{Experimental protocol.}
For each baseline, we identify its set of originally failing queries (ReFoRCE: 638 queries; APEX-SQL: 655 queries). For each failing query, we inspect its D1--D7 scores and inject a targeted natural-language hint into the system's schema context. Hints are dimension-specific: for example, a query with high D4 (multi-stage logic, score $\geq 3$) receives the directive ``\emph{this query requires a multi-stage CTE pipeline; plan the stages before writing SQL}''. %a high-D6 query (temporal precision) receives date-function guidance specific to the DevRev dialect. No changes are made to the baseline agent's core algorithm or generation loop.

\paragraph{Results.}

%\begin{figure}[h]
%  \centering
%  \includegraphics[width=\columnwidth]{%latex/figures/sds_improvement_simple.png}
 % \caption{Overall Answer Correctness before and after SDS-guided hint
 % injection.}
  %(27.2\%~$\to$~41.4\%).}
 % \label{fig:sds_improvement}
%\end{figure}

\begin{table}[h]
\centering
\small
\setlength{\tabcolsep}{4pt}
\begin{tabular}{@{}lrrr@{}}
\toprule
\textbf{Category} & \textbf{ReFoRCE} & \textbf{APEX-SQL} & \textbf{FlexSQL} \\
                  & $\Delta$AC (pp)  & $\Delta$AC (pp)   & $\Delta$AC (pp)  \\
\midrule
Rel./Network                 & +34.9 & +13.9 & +14.3 \\
Classif./Grouping            & +31.1 & +13.6 & +11.5 \\
Date/Time                    & +35.5 & +15.2 & +13.3 \\
Ranking/Top-N                & +39.6 & +12.1 & +13.9 \\
Statistical                  & +33.6 & +12.7 & +17.1 \\
Comparison                   & +36.6 & +15.1 & +16.6 \\
\midrule
\textbf{Overall (all 900)}   & \textbf{+34.1} & \textbf{+14.2} & \textbf{+13.3} \\
\bottomrule
\end{tabular}
\caption{Per-category AC improvement (SDS-guided minus baseline) for ReFoRCE, APEX-SQL, and FlexSQL.}
\label{tab:sds_per_cat}
\end{table}

%Figure~\ref{fig:sds_improvement} reports the results. 
On the 638 originally failing ReFoRCE queries, SDS-guided hints achieve 48.1\% AC on that subset, lifting the score from 29.1\% to 63.2\%. For APEX-SQL's 655 failing queries, hints achieve 19.5\% on that subset, raising AC from 27.2\% to 41.4\%. The larger gain for ReFoRCE reflects the severity of its original dialect gap. %: many baseline queries produced no SQL at all, and the SDS hints that guide it toward the correct analytical structure are sufficient for it to recover on a substantial fraction of previously blank cases.

Table~\ref{tab:sds_per_cat} breaks down the gains by category. Improvements are consistent across all six discriminative categories, with Ranking/Top-N (+39.6~pp for ReFoRCE) and Date/Time (+35.5~pp) showing the largest gains. These categories map directly to SDS dimensions D4 (multi-stage logic) and D6 (temporal precision), confirming that the SDS rubric correctly identifies the loci of difficulty.

The upshot: (I) The SDS rubric is \emph{predictive}: the failing queries that respond most strongly to SDS hints are those with high D4 and D2 (derived metrics) scores. (II) SDS is \emph{actionable}: dimension-score inspection at query time is sufficient to construct targeted guidance that measurably improves accuracy without architectural changes. % This positions SDS not only as a benchmark stratification tool but as a practical component of a human-in-the-loop or automated improvement pipeline for enterprise NL2SQL systems.

\subsection{Ablations}
\label{sec:ablation}

\begin{figure}[t]
\centering
\includegraphics[width=\columnwidth]{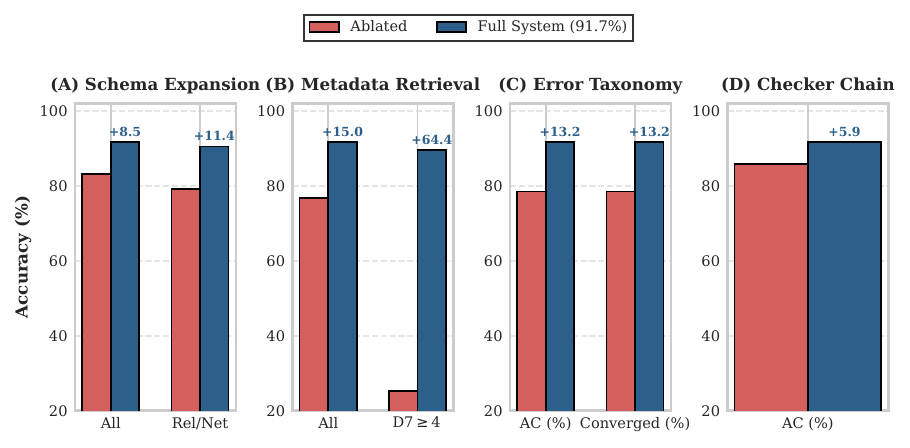}
\caption{Component ablations on the DevRev-900 benchmark. Each panel compares the ablated configuration (red) against the full system at 91.7\% AC (blue).}
\label{fig:ablation}
\end{figure}

%\paragraph{(A) Schema Expansion.} 

\noindent
{\bf (A) Schema Expansion.} We disable Stage~4: on schema-coverage errors (\texttt{table\_not\_found}, \texttt{column\_not\_found}, \texttt{unresolved\_alias}), the orchestrator returns the error to the generator without fetching additional tables. Accuracy drops by 8.5~pp, with a larger 11.4~pp drop on Relationship/Network queries, confirming that polymorphic link traversal is the primary beneficiary of iterative schema expansion.

\noindent
{\bf (B) Metadata-Enriched Retrieval.} We replace Stage~3 with type-signature-only context: column names and DDL types, but no sub-field schemas, enum vocab, sample values, or join hints. This produces the largest ablation effect: $-$15.0~pp overall and a catastrophic $-$64.4~pp on high-D7 queries.
%(those requiring \texttt{LATERAL~FLATTEN} and nested-type constructs). Without sub-field metadata, the generator cannot reason about array structure.

\noindent
{\bf (C) Error Taxonomy.} We replace the structured taxonomy with a generic retry loop: on any failure, the error message is appended and the generator retries without classification, targeted directives, or history preservation. Accuracy drops by 13.2~pp and mean repair rounds increase from 1.4 to 3.2, showing that structured feedback improves convergence rate and reduces wasted iterations.

\noindent
{\bf (D) Deterministic Checkers.} We remove the pre-execution checker chain and send generated SQL directly to the database. Accuracy drops by 5.9~pp, but the cost impact is more significant: mean DB executions per query increase from 1.2 to 2.8. The checkers absorb failures cheaply before they reach the database, reducing both cost and latency.

\section{Conclusion}
\label{sec:conclusion}
% ============================================================================

We present two complementary contributions for NL-to-SQL on nested enterprise schemas. The DevRev benchmark fills a measurement gap, surfacing nested-type traversal, polymorphic link-graph reasoning. %, and persona-aware disambiguation that current benchmarks do not exercise. 
The cost-aware single-generation architecture responds to the regime properties the benchmark exposes, with iterative schema selection, sub-field-aware metadata retrieval, and a structured history-aware error taxonomy.

%On DevRev the architecture attains 91.7\% Answer Correctness overall and 91.5\% on the high-SDS regime (SDS~$\geq 11$, 90.4\% of the corpus). On the Spider~2.0 Snowflake public leaderboard it is competitive with leading systems while using one generation pass per query, occupying a distinct operating point on the cost--accuracy frontier.

The architecture composes naturally with tournament-style multi-candidate generation: in our view the most direct path to extend the work. We also expect the design principles to transfer to other dialects and enterprise platforms whose schemas exhibit the same properties.%, an empirical question we leave to future work.
% ============================================================================
\subsection*{Limitations}
\label{sec:limitations}
% ============================================================================

The system relies on an execution API for the iterative error-feedback loop, and thus cannot run in offline settings. %In offline settings where execution is unavailable, the Stage~4 schema expansion and the runtime portion of the error taxonomy cannot operate; the system reduces to a single-pass generator with deterministic checkers only.
The evaluation covers two benchmarks, both on Snowflake, and generalization to other SQL dialects should be validated. %(PostgreSQL, MySQL, BigQuery) and to enterprise platforms outside the work-item-graph regime remains to be validated.
The DevRev benchmark was generated with LLM assistance and verified by execution against the production database. This is a different verification standard than fully human-curated benchmarks.
%: every query passes a real database execution, but a human did not re-author the natural language.% to confirm that the executed SQL is the most natural interpretation of the question.
Cost comparisons against multi-candidate systems depend on disclosed candidate counts and execution-call counts. For systems whose costs are not publicly reported, our comparisons in Section~\ref{sec:cost} are inferential and treat the candidate count $N$ as a lower bound on inference cost.

\bibliography{custom}

% ============================================================================
\clearpage
\appendix
% ============================================================================

\section{Dataset Release}
\label{app:release}

The DevRev NL2SQL benchmark will be released publicly upon publication. The release includes: (1) all 900 natural-language queries with SDS annotations and complexity-category labels; (2) the full schema documentation (m-schema format) for all 15 entity types used in generation; and (3) the SDS scoring rubric and LLM judge prompt for reproducible evaluation. The dataset will be hosted on Hugging Face under a permissive open-source license. A leaderboard for community submissions will accompany the release.

\section{System Design Details}
\label{app:system-design}
% ============================================================================

\subsection{Offline Knowledge Graph Construction}
\label{app:offline-kg}

A knowledge graph is built once per deployment from the source schema documentation. Three transformations matter for enterprise schemas:

\begin{itemize}[noitemsep,leftmargin=*]
  \item \textbf{Partition Collapsing.} Date-partitioned tables (e.g., \texttt{GA\_SESSIONS\_YYYYMMDD}, \texttt{EVENTS\_YYYYMMDD}) are collapsed into a single logical entry in the catalog, with the partition pattern captured as metadata. This reduces a catalog of millions of physical tables to thousands of logical ones.
  \item \textbf{Key-Column Scoring.} Each table's columns are scored as candidate keys based on uniqueness, naming heuristics, and foreign-key references; the highest-scoring keys are surfaced directly in the catalog to aid table selection without requiring full schema retrieval.
  \item \textbf{Type Normalization.} Source-specific type strings are mapped to a canonical type vocabulary (\texttt{ARRAY[STRUCT]}, \texttt{STRUCT}, primitive types) so that downstream consumers reason about nested structure uniformly.
\end{itemize}

\subsection{Error Taxonomy Classes}
\label{app:error-taxonomy}

When generation produces SQL that fails on execution, the orchestrator classifies the failure into one of 15+ error types. The taxonomy is organized by failure phase, with each class mapped to a corrective action:

\begin{itemize}[noitemsep,leftmargin=*]
  \item \textbf{Schema-coverage errors:} \texttt{table\_not\_found}, \texttt{column\_not\_found}, \texttt{unresolved\_alias}. Trigger Stage~4 schema expansion.
  \item \textbf{Nested-type errors:} \texttt{forbidden\_lateral} (incorrect \texttt{LATERAL} placement), \texttt{unnest\_single\_list} (flattening a singleton array), \texttt{nested\_conversion\_error} (type coercion over a sub-field). Trigger repair with a construct-specific directive.
  \item \textbf{Identifier errors:} \texttt{reserved\_word\_alias} (alias collides with a Snowflake reserved word), \texttt{ambiguous\_flatten\_alias} (multiple flattens with conflicting aliases). Trigger alias-rewriting repair.
  \item \textbf{Semantic errors:} \texttt{empty\_result}, \texttt{wrong\_aggregation\_grain}, \texttt{missing\_filter}. Trigger semantic repair with the question re-presented alongside the failed SQL.
\end{itemize}

\subsection{Deterministic Checker Chain}
\label{app:checker-chain}

Table~\ref{tab:checker-chain} lists all ten stages of the deterministic checker chain (Section~\ref{sec:checkers}). Stages are executed in order; blockers reject the SQL and return classified errors to the orchestrator's repair loop, while rewriters silently transform the SQL before passing it to the next stage. Execution against the database is the final stage, ensuring that all statically detectable failures are absorbed before incurring database cost.

\begin{table}[t]
\centering
\scriptsize
\begin{tabularx}{\columnwidth}{@{}llX@{}}
\toprule
\textbf{Checker} & \textbf{Type} & \textbf{Purpose} \\
\midrule
Multi-Statement & Blocker & Rejects multiple statements \\
\texttt{SELECT *} Guard & Blocker & Rejects \texttt{SELECT *} patterns \\
\texttt{VALUES} Guard & Blocker & Rejects \texttt{VALUES(...)} in CTEs \\
Placeholder Guard & Blocker & Rejects bind parameters/stubs \\
Dialect Rewriter & Rewriter & Silently fixes dialect mistakes \\
Reserved Word Guard & Blocker & Rejects reserved word aliases \\
Syntax Validator & Blocker & Local parse check (\texttt{sqlglot}) \\
Fuzzy Search Guard & Blocker & Rejects \texttt{LIKE}/\texttt{ILIKE} patterns \\
Heuristic Detector & Blocker & Regex checks for anti-patterns \\
Execution Validator & Blocker & Executes against the database \\
\bottomrule
\end{tabularx}
\caption{Deterministic Checker Chain. Blockers reject queries and return errors to the orchestrator; Rewriters silently mutate the SQL.}
\label{tab:checker-chain}
\end{table}

% ============================================================================
\section{Dataset Generation Pipeline: Full Details}
\label{app:pipeline}
% ============================================================================

This appendix provides the complete specification of the four-stage
formula-anchored generation pipeline introduced in
Section~\ref{sec:pipeline}, together with the full SDS scoring rubric
(Section~\ref{app:sds}), the LLM judge rubric used for evaluation
(Section~\ref{app:judge}), and annotated examples of each intermediate
artifact (Section~\ref{app:examples}).

% ----------------------------------------------------------------------------
\subsection{Stage 0: Sub-Schema Generation}
\label{app:stage0}
% ----------------------------------------------------------------------------

Before the four-stage query pipeline begins, an \emph{LLM sub-schema
generator} produces a semantically coherent column subset from each of
the 14 table combinations (e.g., \texttt{account + enhancement + ticket},
\texttt{account + meeting + opportunity}).  The generator receives the
full \texttt{MSCHEMA} source for the relevant tables, a structured
text format that records, for each field, its type, description, enum
vocabulary, and sub-field schema if the column is an
\texttt{ARRAY[STRUCT]}, and is instructed to select 2--5
analytically useful non-connection columns per table while automatically
retaining all primary-key and foreign-key columns.  The output is a JSON
sub-schema (\texttt{sub\_schema.\{table\}.\{columns\}}) that supplies
Stage~1 with a pruned, focused view of the schema.  Across 14 table
combinations, 150 sub-schemas are produced, forming the atomic unit of
the pipeline.

% ----------------------------------------------------------------------------
\subsection{The Formula Book}
\label{app:formula-book}
% ----------------------------------------------------------------------------

A \emph{formula book} of 25 structural patterns is derived from the 25
highest-SDS Spider~2.0-Snow queries (all with SDS~$\geq 27$ on the
0--35 scale).  Each formula is stored in two representations.

\paragraph{Human-readable representation.}
Each entry contains: (i)~the source Spider~2.0 query text; (ii)~a
\textbf{Level~1 macro pattern}: a multi-stage logic skeleton expressed
in slot-variable notation; and (iii)~\textbf{Level~2 micro patterns}: a
list of semantic-depth phrases from the source query, each annotated with
the SDS dimension it triggers and the SQL pattern it forces.

\smallskip
\noindent\textbf{Example: Formula F\_001 (snapshot eligibility + percentage change).}

\medskip
\noindent\textit{Source query (abbreviated):} ``For each U.S.\ state, find how the number of
active financial branch entities has changed from March~1, 2020 to
December~31, 2021.  An entity is considered active on a specific date if
its start date is on or before that date and its end date is either null
or on or after that date.''

\medskip
\noindent\textit{Macro pattern (Level~1):}
\begin{enumerate}[noitemsep,leftmargin=*]
  \item Define \textsc{[concept]} using a plain-English eligibility rule:
    \texttt{start\_date}~$\leq$~date \textsc{and}
    (\texttt{end\_date}~\textsc{is null} \textsc{or}
     \texttt{end\_date}~$\geq$~date).
  \item Evaluate \textsc{[concept]} at \textsc{[snapshot\_date\_1]} for
    each \textsc{[grouping\_dimension]}.  Evaluate at
    \textsc{[snapshot\_date\_2]}.
  \item Compute \textsc{[derived\_metric: \% change]} between snapshots
    per \textsc{[grouping\_dimension]}.
  \item Output: grouping, snapshot-1 value, snapshot-2 value, \%~change.
\end{enumerate}

\noindent\textit{Micro patterns (Level~2, selected):}
\begin{itemize}[noitemsep,leftmargin=*]
  \item ``A \textsc{[entity]} is considered \textsc{[concept]} on a
    specific date if \ldots and its end date is either null or on or
    after that date'' $\to$ \textbf{D1} trigger (multi-condition NULL
    eligibility).
  \item ``the percentage change in these counts'' $\to$ \textbf{D2}
    trigger (derived metric = $(v_2 - v_1)/v_1 \times 100$).
  \item ``find how \textsc{[metric]} has changed from \textsc{[date\_1]}
    to \textsc{[date\_2]}'' $\to$ \textbf{D6} trigger (two precise
    snapshot dates).
\end{itemize}

\noindent\textit{Filled example on DevRev schema:} ``For each account tier, find how the number of actively open tickets changed between July~1, 2023, and December~31, 2023.  A ticket counts as actively open on a given date if it was created on or before that date and its resolution date is either missing or falls after that date.  For each tier, show the open count on July~1st, the open count on December~31st, and the percentage change, sorted by the largest drop.''

\paragraph{Machine-readable representation.}
The same 25 formulas are stored as \texttt{SPIDER2\_FORMULA\_BOOK.json},
with fields: \texttt{formula\_id}, \texttt{sds\_score},
\texttt{sds\_dimensions} (per-dimension scores), \texttt{complexity\_categories},
\texttt{macro\_pattern}, \texttt{micro\_patterns} (list), \texttt{filled\_example},
\texttt{slot\_type\_categories} (lists of required column types: date,
categorical, numeric, geographic), and
\texttt{recommended\_for\_slots} (complexity-category tags used by
Stage~2 to score compatibility).

% ----------------------------------------------------------------------------
\subsection{Stage 1: Six-Slot Query Planning}
\label{app:stage1}
% ----------------------------------------------------------------------------

Stage~1 issues a single LLM call per sub-schema and produces a
\textbf{six-slot plan} (\texttt{plan.json}).  Each slot specifies the
analytical work that one of the six generated queries must do.

\paragraph{Slot fields.} Each slot in \texttt{plan.json} contains:
\begin{itemize}[noitemsep,leftmargin=*]
  \item \texttt{slot\_number} (1--6) and \texttt{length\_slot}:
    \texttt{SHORT}~(15--25 words), \texttt{SHORT-MEDIUM}~(30--50),
    \texttt{MEDIUM}~(50--80), \texttt{MEDIUM-LONG}~(80--120),
    \texttt{LONG}~(120--150), \texttt{VERY-LONG}~(150--175).
  \item \texttt{complexity\_categories}: subset of the eight analytical
    categories (Aggregation, Relationship/Network, Date/Time, etc.).
  \item \texttt{sds\_target}: per-dimension targets D1--D7, expressed
    on a 0--2 scale (planning scale; scoring uses 0--5).
  \item \texttt{columns\_to\_use}: mapping of table~$\to$~column list
    for this slot, drawn from the sub-schema.
  \item \texttt{analytical\_intent}: one-sentence description of the
    computation the slot must express.
  \item \texttt{required\_formula\_patterns}: a list of structural
    keywords (e.g., \texttt{two\_stage\_drill}, \texttt{cohort\_retention})
    used by Stage~2 to score formula compatibility.
\end{itemize}

\paragraph{Global coverage constraint.}
The Stage~1 prompt enforces minimum category coverage across all six
slots: Aggregation in all six; Relationship/Network in at least five;
Classification/Grouping in at least four; Date/Time in at least three;
Ranking/Top-N, Statistical, and Comparison each in at least two.  A
\texttt{distribution\_check} field in the plan output surfaces whether
these constraints were satisfied; plans that violate any constraint are
discarded and regenerated.

% ----------------------------------------------------------------------------
\subsection{Stage 2: Formula Retrieval (Programmatic)}
\label{app:stage2}
% ----------------------------------------------------------------------------

Stage~2 is a pure Python program with no LLM calls.  For each slot
it scores every formula in \texttt{SPIDER2\_FORMULA\_BOOK.json} on two
axes and assembles an ordered list:

\begin{enumerate}[noitemsep,leftmargin=*]
  \item \textbf{Category overlap.}  The Jaccard similarity between the
    slot's \texttt{complexity\_categories} and the formula's
    \texttt{complexity\_categories}.
  \item \textbf{Pattern compatibility.}  Whether the slot's
    \texttt{required\_formula\_patterns} intersect the formula's
    macro-pattern keywords, and whether the formula's required column
    types (\texttt{slot\_type\_categories}) are satisfied by the columns
    available in the slot's sub-schema.
\end{enumerate}

Up to five formulas per slot are retained (\texttt{samples\_per\_category~=~5}).
A global no-repeat constraint ensures no formula is assigned to more
than one slot within a sub-schema.  The output is
\texttt{formula\_examples.json}, keyed by slot, with each entry
containing \texttt{formula\_id}, \texttt{macro\_pattern},
\texttt{micro\_patterns}, and \texttt{filled\_example} (SDS scores are
stripped to prevent the generator from over-fitting to them).

% ----------------------------------------------------------------------------
\subsection{Stage 3: Per-Slot Query Generation}
\label{app:stage3}
% ----------------------------------------------------------------------------

Given the slot plan and the retrieved formula examples, Stage~3 issues
one LLM call per slot.  The prompt provides: the slot's sub-schema
(column names, types, descriptions, enum vocabularies), the
\texttt{analytical\_intent} and \texttt{required\_formula\_patterns} from
the plan, and the formula macro-pattern plus a filled example from
Stage~2.  The filled example is instantiated on a \emph{different}
schema to provide structural guidance without leaking DevRev vocabulary.

\paragraph{Per-slot output (\texttt{slot\_queries\_from\_examples.json}).}
Each slot record contains:
\begin{itemize}[noitemsep,leftmargin=*]
  \item \texttt{column\_mapping}: each column labeled with its
    role: \texttt{GROUPING}, \texttt{METRIC}, \texttt{FILTER},
    \texttt{OUTPUT}, or \texttt{ANCHOR}.
  \item \texttt{query\_plan}: a 2--5 sentence structural blueprint
    (``Scope: \ldots; Stage~1: \ldots; Stage~2: \ldots; Output: \ldots'').
  \item \texttt{query}: the natural-language query text, bounded by the
    slot's target word count.
  \item \texttt{formula\_ids\_used}, \texttt{categories\_used},
    \texttt{reasoning}: provenance metadata.
\end{itemize}

\noindent All generated queries are executed against the DevRev
database; queries returning empty results are discarded (7.6\%
rejection rate across the full corpus).

% ----------------------------------------------------------------------------
\subsection{Stage 4: Persona Rewriting}
\label{app:stage4}
% ----------------------------------------------------------------------------

Stage~4 issues a single LLM call that receives all six slot queries
simultaneously and rewrites each into a distinct business-role voice.
The six personas are: \textbf{VP of Sales}, \textbf{Customer Success
Manager}, \textbf{Product Manager}, \textbf{Operations Analyst},
\textbf{Finance / Revenue Ops}, and \textbf{C-Suite Executive}.  Each
persona is used exactly once per sub-schema.

\paragraph{Key rewriting rules.}
\begin{itemize}[noitemsep,leftmargin=*]
  \item \textbf{Enum-value ban.}  Categorical field values (status
    enums, segment labels, milestone names) must not appear in the
    rewritten query.  The rewrite names the dimension
    (``by account tier'') without listing its values.  The only
    exception is a single-value filter that is the analytical focus of
    the entire query.
  \item \textbf{Structural diversity.}  Opening styles are varied
    across slots: question, imperative, declarative need, narrative
    framing, concern.
  \item \textbf{Preservation invariants.}  The following properties
    are protected and must survive rewriting: exact date ranges and
    temporal grain; every status and eligibility filter; all numeric
    thresholds; aggregation grain (per-account vs.\ per-category);
    ratio denominators; sort order and ranking direction.
  \item \textbf{Interpretability test.}  A rewrite is valid if and
    only if the original query remains a natural interpretation of it:
    a SQL agent reading only the rewrite must be able to produce the
    same SQL as the original.
\end{itemize}

\paragraph{Self-check protocol.}
The Stage~4 prompt includes a mandatory 15-item pre-return checklist
covering: enum-value scan, time-window preservation, temporal grain,
filter completeness, aggregation grain, ratio denominator, sort
direction, and absence of SQL keywords in the output.

\paragraph{Quality audit.}
An automated column-plan audit verifies that 100\% of rewritten queries
pass their SDS slot threshold (as scored by the SDS rubric in Section~\ref{app:sds}) and that
76.9\% achieve exact column-plan match (i.e., the query implies the
same columns as the original plan without additional retrieval).

% ----------------------------------------------------------------------------
\subsection{SDS Scoring Rubric (Full Specification)}
\label{app:sds}
% ----------------------------------------------------------------------------

The Semantic Depth Score (SDS) is assigned by an LLM judge that receives
the natural-language query text only (no SQL and no schema).  Each of
the seven dimensions is scored 0--5 (max total: 35).  The rubric
includes calibration examples anchoring Easy~$\approx 7$,
Medium~$\approx 15$--18, and Hard~$\approx 29$.  The complete rubric is
reproduced below.

\begin{table*}[t]
\centering\small
\begin{tabular}{@{}ll p{\dimexpr\textwidth-6.5cm\relax}@{}}
\toprule
\textbf{Dim.} & \textbf{Name} & \textbf{What a score of 5 requires} \\
\midrule
D1 & Eligibility            & Rank/percentile-based, anti-join, or recursive set-logic eligibility (e.g., groups whose ticket volume grew by at least 10\% QoQ, excluding the bottom quartile for enhancement-linked revenue). \\[2pt]
D2 & Derived Metric         & Cascaded derivation: a formula whose inputs are themselves derived metrics (e.g., percentile rank of a ratio, running total expressed as \% of grand total, or ratio of two computed rates). \\[2pt]
D3 & Population Scoping     & Nested scope: the eligible population is defined by the result of an inner aggregation or ranking (e.g., enhancements whose linked ticket count places them in the top decile of ticket volume). \\[2pt]
D4 & Multi-Stage Logic      & Three or more interdependent CTEs, a recursive structure, or multiple fully independent aggregations joined together (e.g., a CTE chain that computes quarterly totals, per-partner normalized counts, intersects both ranked lists, then breaks by month). \\[2pt]
D5 & Output Precision       & Output broken out across two grouping dimensions simultaneously, with computed fields for each cell (e.g., for each group $\times$ month combination, show ticket count and revenue share). \\[2pt]
D6 & Temporal Precision     & Multi-dimensional temporal logic: cross-period comparison with sub-period breakdown on both axes, or entity-relative windows combined with calendar periods. \\[2pt]
D7 & SQL Construct Diversity & Full complexity: three or more of the following required together: recursive CTE, multiple window functions, set operations (\texttt{UNION}/\texttt{INTERSECT}), anti-join, conditional aggregation, dependent CTE chain. \\
\bottomrule
\end{tabular}
\caption{SDS dimension definitions at maximum score (5/5). Lower scores are proportionally less complex; a score of 0 indicates the dimension is entirely absent from the query.}
\label{tab:sds_rubric}
\end{table*}

\paragraph{Score-5 anchor examples (abbreviated).}
\begin{itemize}[noitemsep,leftmargin=*]
  \item \textbf{D1~=~5:} ``Groups whose ticket volume grew by at least
    10\% quarter-over-quarter, \emph{excluding} any group that also
    appears in the bottom quartile for enhancement-linked revenue.''
  \item \textbf{D4~=~5:} ``A CTE chain that (1) computes quarterly
    totals, (2) computes per-partner normalized counts, (3) intersects
    both ranked lists, and (4) breaks results by month within each
    quarter.''
  \item \textbf{D7~=~5:} ``Two window functions + correlated subquery +
    conditional aggregation + three-CTE chain.''
\end{itemize}

\paragraph{Scoring guidance.}
The rubric instructs the scorer to award scores of 3 or higher only
when specific phrases in the query text justify the level, and to
resist inflating scores for query length or business jargon.  The
cross-check step compares the candidate total against the Hard anchor
($\approx 29$) and requires downward revision if the candidate total
exceeds it for a query of lesser complexity.

% ----------------------------------------------------------------------------
\subsection{LLM Judge Rubric}
\label{app:judge}
% ----------------------------------------------------------------------------

The LLM judge used for evaluation (Section~\ref{sec:setup}) scores each
generated SQL on a maximum of 50 points across four criteria.

\begin{table*}[t]
\centering\small
\begin{tabular}{@{}llr p{\dimexpr\textwidth-7cm\relax}@{}}
\toprule
\textbf{Dim.} & \textbf{Name} & \textbf{Max} & \textbf{Focus} \\
\midrule
J1 & Semantic Completeness & 35 & Does the SQL answer every part of the
     question?  Hard caps apply if one or more requirements are entirely
     absent. \\
J2 & Analytical Logic      & 8  & Are CTEs, window functions,
     percentage-change formulas, and multi-stage logic correctly
     implemented? \\
J3 & Tables \& Joins       & 4  & Are the correct tables joined via the
     right keys without spurious cross-products? \\
J4 & Output Precision      & 3  & Are the output columns correctly named,
     typed, and ordered as the question specifies? \\
\midrule
    & \textbf{Total}       & \textbf{50} & \\
\bottomrule
\end{tabular}
\caption{LLM judge rubric dimensions (max 50 points).
  \textbf{Answer Correctness (AC)} is the binary verdict derived from
  this rubric: AC~=~1 if and only if the judge's overall assessment is
  that the SQL semantically answers the question.}
\label{tab:judge_rubric}
\end{table*}

\paragraph{Evaluation protocol.}
The judge receives: the natural-language question, the generated SQL,
the ground-truth schema (DDL), and a preview of the first five result
rows.  Before scoring, the judge completes a mandatory checklist:
entity scope, filter conditions, grouping keys, metric definitions,
ranking directives, output columns, multi-stage structure, and
comparison direction.  Each checklist item is mapped to a specific
SQL clause before a score is assigned.  A J1 hard cap is applied if
one or more checklist items are entirely unaddressed.

% ----------------------------------------------------------------------------
\subsection{Annotated Pipeline Examples}
\label{app:examples}
% ----------------------------------------------------------------------------

Table~\ref{tab:pipeline_trace} traces one instance
(\texttt{account\_enhancement\_ticket\_1}, Slot~6) through all four
stages to illustrate how a formula macro-pattern becomes a final
persona-rewritten query.

\begin{table*}[t]
\centering\small
\begin{tabular}{@{}l p{\dimexpr\textwidth-3.5cm\relax}@{}}
\toprule
\textbf{Stage} & \textbf{Artifact} \\
\midrule
\textbf{Stage 1} \newline (Plan)
  & \textit{Slot:} 6 \quad \textit{Length:} VERY-LONG (150--175 words) \newline
    \textit{Categories:} Aggregation, Relationship/Network, Date/Time,
    Ranking/Top-N, Statistical \newline
    \textit{SDS targets:} D1=2, D2=2, D3=2, D4=2, D5=2, D6=2, D7=2
    (total 14 on planning scale) \newline
    \textit{Intent:} Identify the top enhancements by Q2 volume
    normalized by active-partner count, restricted to groups with
    $\geq$25\% QoQ growth, with per-account breakdowns. \\
\midrule
\textbf{Stage 2} \newline (Formula)
  & \textit{Formula assigned:} F\_002 (find-winner-then-re-aggregate
    at multiple granularities) \newline
    \textit{Macro pattern:} Stage~1 finds the winning dimension by a
    metric; Stage~2 re-aggregates for that winner at finer granularity;
    Stage~3 applies a unit conversion and outputs the maximum per
    granularity. \\
\midrule
\textbf{Stage 3} \newline (Raw query)
  & ``For Q1 and Q2~2025, only include ticket ownership groups where
    the number of tickets rose by at least 25\% from Q1 to Q2.
    Within those qualifying groups, rank enhancements by their Q2
    volume of linked tickets normalized by the number of active partners
    selling into the associated accounts, and return the top~10.
    For each of those top enhancements, report the theme they belong to
    and include the enhancement's short description.
    Also break the results out by account name: for each linked account,
    show its Q2 rate of tickets per \$1M of annual revenue and its Q2
    average of tickets per active partner for that enhancement.
    Order by the largest Q2 normalized volume; within each enhancement
    list accounts from highest to lowest tickets per \$1M revenue.'' \\
\midrule
\textbf{Stage 4} \newline (Persona rewrite)
  & \textit{Persona:} C-Suite Executive \newline
    ``For the exec briefing and midyear planning (using tickets'
    creation dates owned by each group), compare Q1~2025 to Q2~2025
    and only keep ticket ownership groups whose ticket count rose by at
    least 25\% from Q1 to Q2.  Within those qualifying groups, rank
    enhancements by their Q2 linked-ticket volume normalized by the
    number of active partners selling into the associated accounts, and
    return the top~10 enhancements.  For each, include its theme and a
    short description, and break results out by linked account: for each
    account show its Q2 rate of tickets per \$1M of annual revenue and
    its Q2 average of tickets per active partner for that enhancement.
    Order enhancements by the largest Q2 normalized volume; within each
    enhancement list accounts from highest to lowest by tickets per
    \$1M revenue.'' \\
\bottomrule
\end{tabular}
\caption{End-to-end pipeline trace for one instance
  (\texttt{account\_enhancement\_ticket\_1}, Slot~6), showing how the
  Stage~1 plan and Stage~2 formula combine to produce the Stage~3 raw
  query, which Stage~4 then rewrites into a C-Suite voice.}
\label{tab:pipeline_trace}
\end{table*}

% ============================================================================
\section{Additional Experimental Results}
\label{app:additional}
% ============================================================================
\subsection{Results on DevRev: Detailed Breakdown}
\label{sec:devrev-results}

The system attains 91.7\% Answer Correctness (825/900 queries) on the full 900-query DevRev benchmark, and 91.5\% (745/814) on the high-SDS subset (SDS~$\geq 11$, 90.4\% of the corpus). Table~\ref{tab:devrev_headline} shows the cumulative contribution of each component.

\begin{table}[h]
\centering
\small
\begin{tabular}{lr}
\toprule
\textbf{System} & \textbf{AC (\%)} \\
\midrule
Single-pass generator (no pipeline) & 38.2 \\
\quad $+$ schema selection (Stages 1--3) & 46.6 \\
\quad $+$ error taxonomy & 86.8 \\
\quad $+$ Stage 4 expansion (full system) & \textbf{91.7} \\
\bottomrule
\end{tabular}
\caption{Cumulative component contribution on DevRev-900.}
\label{tab:devrev_headline}
\end{table}

Table~\ref{tab:devrev_sds} stratifies accuracy by queries that score at the maximum (5) on each SDS dimension. Failures on high-D2 queries (derived metrics) implicate the generator's reasoning over computed values; failures on high-D6 queries (temporal precision) implicate date-function handling; failures on high-D7 queries (advanced constructs including \texttt{LATERAL~FLATTEN}) implicate the nested-type machinery the architecture is designed for.

\begin{table}[h]
\centering
\small
\begin{tabular}{lrr}
\toprule
\textbf{Subset} & \textbf{Queries} & \textbf{AC (\%)} \\
\midrule
All                        & 900 & 91.7 \\
D1 $=5$ (eligibility)      & 277 & 93.5 \\
D2 $=5$ (derived metric)   &  26 & 84.6 \\
D3 $=5$ (population)       & 122 & 90.2 \\
D4 $=5$ (multi-stage)      & 136 & 94.9 \\
D5 $=5$ (output precision) &  96 & 95.8 \\
D6 $=5$ (temporal)         &  22 & 90.9 \\
D7 $=5$ (constructs)       &  24 & 91.7 \\
\bottomrule
\end{tabular}
\caption{DevRev accuracy stratified by SDS dimension at maximum score (5 on the 0--5 scale).}
\label{tab:devrev_sds}
\end{table}

\subsection{SDS-Stratified Per-Bucket Breakdown}
\label{app:sds_breakdown}

Table~\ref{tab:sds_full} reports mean judge score for all four systems across the three SDS buckets. All three baselines decline monotonically as SDS increases, while our system remains within a 2.8~pp range.

\begin{table*}[h]
\centering\small
\setlength{\tabcolsep}{10pt}
\begin{tabular}{lcccc}
\toprule
\textbf{SDS Bucket} & \textbf{Ours} & \textbf{APEX-SQL} & \textbf{FlexSQL} & \textbf{ReFoRCE} \\
\midrule
Easy ($\leq$12, $n$=139)   & 80.1 & 56.5 & 55.1 & 48.5 \\
Medium (13--18, $n$=250)   & 80.9 & 53.7 & 48.9 & 36.2 \\
Hard ($>$18, $n$=511)      & 78.1 & 41.4 & 42.8 & 27.2 \\
\midrule
Drop (Easy $\to$ Hard)     & \phantom{0}2.0 & 15.1 & 12.3 & 21.3 \\
\bottomrule
\end{tabular}
\caption{Mean judge score (\% of 50-point max) by SDS bucket. Corresponding Answer Correctness figures for our system: 89.2\% (Easy), 94.4\% (Medium), 91.0\% (Hard).}
\label{tab:sds_full}
\end{table*}

\subsection{Query Complexity Categories}
\label{sec:categories}

Each generated query is tagged with one or more \emph{complexity categories} that identify the kind of analytical reasoning the query requires. We distinguish two orthogonal axes.

\textbf{Eight analytical categories} capture the type of computation the SQL must perform:

\begin{itemize}[noitemsep,leftmargin=*]
  \item \textbf{Aggregation.} Summarizing raw records into a computed value. The most universal category; enforced in all six generation slots.
  \item \textbf{Statistical.} Distributional statistics (std deviation, median, percentile, growth rate). In DevRev, this covers CSAT score distributions, ticket-volume trends, and resolution-time dispersion.
  \item \textbf{Comparison.} Contrasting two groups, two time periods, or two metrics (``Q1 vs.\ Q2,'' ``Customer vs.\ Partner tickets,'' ``before and after escalation''). Requires conditional aggregation or a self-join, and is a primary source of multi-stage logic.
  \item \textbf{Classification/Grouping.} Partitioning entities into labeled groups defined by a categorical column or a stated rule. Tests \texttt{GROUP BY} on the right dimension, including over enum-constrained string columns.
  \item \textbf{Ranking/Top N.} Top or bottom $N$ entities by a metric, or the $N$th-ranked entity. Requires \texttt{ORDER BY ... LIMIT} or a window function.
  \item \textbf{Date/Time.} A specific temporal scope, time difference, or date arithmetic (rolling windows, quarter boundaries, age-in-days). Tests translation of natural-language time expressions into correct Snowflake date functions.
  \item \textbf{Relationship/Network.} Multi-hop traversal: linking tickets to enhancements via the typed-link graph, joining accounts to their opportunities and conversations, navigating group-to-vista hierarchies. Characteristic of the work-item graph regime.
  \item \textbf{Geographic.} Filtering or grouping by a geographic dimension. Applied only when a geographic column is present in the sub-schema; rare in DevRev (a B2B SaaS product without location-bearing entities).
\end{itemize}

\paragraph{Domain categories} identify which DevRev business objects the query is primarily about: Ticket/Support, Account/Customer, Opportunity/Sales, Product/Feature, Meeting/Engagement, Dashboard, and so on. These categories are not enforced by the generation pipeline; they emerge from the table combination selected, and are useful for downstream per-domain failure-rate analysis.

\paragraph{Enforced coverage.}
Stage~1 enforces minimum analytical-category coverage across the six slots of every sub-schema: Aggregation in all six; Relationship/Network in at least five; Classification/Grouping in at least four; Date/Time in at least three; Ranking/Top-N, Statistical, and Comparison each in at least two. This prevents the corpus from collapsing to a single query type and ensures every table combination exercises the full range of analytical reasoning patterns.

Table~\ref{tab:categories} shows the per-query distribution across the final 900-query corpus.

\begin{table}[h]
\centering
\small
\begin{tabular}{llr}
\toprule
\textbf{Category} & \textbf{Queries (of 900)} & \textbf{\%} \\
\midrule
Aggregation              & 900 & 100.0 \\
Relationship/Network     & 719 &  79.9 \\
Classification/Grouping  & 705 &  78.3 \\
Date/Time                & 594 &  66.0 \\
Ranking/Top N            & 323 &  35.9 \\
Statistical              & 292 &  32.4 \\
Comparison               & 205 &  22.8 \\
Geographic               &   1 &   0.1 \\
\bottomrule
\end{tabular}
\caption{Per-query complexity-category distribution. Each query is assigned multiple categories; percentages reflect the fraction of all 900 queries that include each category. Aggregation is universal by design.}
\label{tab:categories}
\end{table}

\subsection{Dataset Statistics}
\label{sec:stats}

The pipeline produces 900 execution-verified natural-language queries over 14 schema combinations. Table~\ref{tab:dataset_stats} compares corpus-level statistics for DevRev and Spider~2.0-Snow.

\begin{table}[h]
\centering
\small
\begin{tabular}{lrr}
\toprule
\textbf{Metric} & \textbf{DevRev} & \textbf{Spider~2.0-Snow} \\
\midrule
Queries & 900 & 547 \\
Mean SDS & 19.15 & 20.20 \\
Avg.\ word count & 29.3 & 54.0 \\
Self-BLEU & 0.039 & 0.023 \\
Execution pass rate & 92.4\% & N/A \\
\bottomrule
\end{tabular}
\caption{Corpus-level statistics for DevRev and Spider~2.0-Snow. Self-BLEU measures $n$-gram redundancy (lower is more diverse).}
\label{tab:dataset_stats}
\end{table}

When scored on the same 0--5-per-dimension rubric, DevRev and Spider~2.0-Snow exhibit comparable analytical depth (mean SDS 19.15 vs.\ 20.20). The corpora differ in their \emph{difficulty profile} rather than their average depth. Spider~2.0-Snow concentrates difficulty in SQL-construct diversity (D7: 3.67 vs.\ 2.57) and multi-stage pipeline logic (D4: 3.12 vs.\ 2.72), reflecting its cross-domain schema variety and advanced SQL patterns. DevRev concentrates difficulty in derived-metric construction (D2: 2.92 vs.\ 2.65), output-specification precision (D5: 3.45 vs.\ 2.88), and temporal reasoning (D6: 1.51 vs.\ 1.33), reflecting the CRM domain where percentage-change comparisons and rolling-window time filters are standard. The self-BLEU of 0.039 confirms lexical diversity is maintained despite all queries sharing DevRev domain vocabulary.

% ----------------------------------------------------------------------------
\subsection{Results on Spider~2.0-Snow}
\label{sec:spider-results}
% ----------------------------------------------------------------------------

Spider~2.0-Snow comprises 547 queries, but ground-truth SQL and expected results are publicly available for only 120 of them~\citep{spider2}. We evaluate our system on this 120-query gold-answer subset and report that score alongside the full-leaderboard figures for the three baselines. Table~\ref{tab:spider_leaderboard} shows the results.

\begin{table}[h]
\centering
\small
\begin{tabular}{p{4.2cm}rr}
\toprule
\textbf{System} & \textbf{EX (\%)} & \textbf{Eval.} \\
\midrule
Genloop Sentinel v2 Pro    & 96.70 & 547 \\
Native mini                & 96.53 & 547 \\
QUVI-3 + Gemini-3-pro      & 94.15 & 547 \\
TCDataAgent-SQL            & 93.97 & 547 \\
$\ldots$                   &       &     \\
\textbf{Ours (single gen.)} & \textbf{85.8} {\small(103/120)} & 120$^\dagger$ \\
$\ldots$                   &       &     \\
APEX-SQL \citep{apexsql}   & 73.13 & 547 \\
FlexSQL \citep{flexsql}    & 65.45 & 547 \\
ReFoRCE \citep{reforce}    & 62.89 & 547 \\
\bottomrule
\end{tabular}
\caption{Spider~2.0-Snow leaderboard snapshot (May 2026).
EX = Execution Accuracy. Leaderboard entries for all other systems are evaluated on the full 547 queries.
$^\dagger$Our system is evaluated on the 120-query gold-answer subset for which ground-truth SQL and results are publicly available; the score is therefore not directly comparable to full-leaderboard figures and is not submitted to the ranked leaderboard.
Baseline entries are each system's best leaderboard submission: APEX-SQL (rank~17, default config), FlexSQL (rank~22, with \texttt{gpt-oss-120b}), ReFoRCE (rank~24, with \texttt{o3}).}
\label{tab:spider_leaderboard}
\end{table}

Our system answers 103 of 120 gold-answer queries correctly (85.8\% EX). Because this evaluation covers only the publicly released subset rather than the full 547-query leaderboard set, the figure is indicative rather than directly ranked. The three baselines that score 27--37\% AC on DevRev score 63--73\% EX on Spider~2.0-Snow's full leaderboard, a regime they were optimised for. This contrast confirms that the performance gap on DevRev reflects the structural properties of the enterprise nested-schema regime rather than a general weakness in those systems. Our 85.8\% on the gold subset suggests the architecture is competitive on the flat-table, large-scale schema-linking regime while using one generation pass per query.

% ----------------------------------------------------------------------------
\subsection{Cost Comparison}
\label{sec:cost}
% ----------------------------------------------------------------------------

Table~\ref{tab:cost} compares LLM API calls and database executions per query across all four systems, measured on the same stratified 50-query cost sample. Counts are medians; LLM calls are measured from API-call logs and DB executions from per-query execution traces.

\begin{table}[h]
\centering
\small
\begin{tabular}{p{3.0cm}rrr}
\toprule
\textbf{System} & \textbf{AC (\%)} & \textbf{LLM/q} & \textbf{DB/q} \\
\midrule
APEX-SQL     & 27.2 & 16 & 18 \\
FlexSQL      & 37.1 & 46 & 20 \\
ReFoRCE      & 29.1 &  4 &  4 \\
\textbf{Ours}& \textbf{91.7} & \textbf{2} & \textbf{4} \\
\bottomrule
\end{tabular}
\caption{Cost comparison on the 50-query stratified sample (median values). LLM/q = LLM API calls per query; DB/q = database executions per query. AC = Answer Correctness on the full 900-query DevRev benchmark.}
\label{tab:cost}
\end{table}

Our system matches ReFoRCE on DB executions (median 4) while achieving 3$\times$ higher AC, and uses $8\text{--}23\times$ fewer LLM calls than APEX-SQL and FlexSQL. The multi-candidate baselines incur high DB costs because each of their $K$ candidate plans independently probes the schema with exploratory sub-queries before generating final SQL; our architecture avoids this through a single schema-selection pass guided by the metadata-enriched retriever.

APEX-SQL's hypothesis-verification loop issues a mean of 16 LLM calls per query (median 16), each paired with a DB execution to verify or refute a schema hypothesis, yielding 18 DB calls in the median case. FlexSQL generates 4 candidate plans, each of which independently executes sub-queries per involved table to ground column values before SQL synthesis, resulting in a median of 20 DB executions and 46 LLM calls. ReFoRCE's self-refinement loop (up to 5 iterations) issues one LLM call and one DB execution per iteration, matching our system's DB footprint despite using 2$\times$ more LLM calls. Our system issues a single LLM call for SQL generation; additional calls occur only when the pre-execution checker or execution failure triggers a repair round (median 2 calls total, max 3 in this sample).
\end{document}